\documentclass{article}

\usepackage[preprint]{neurips_2024}
\usepackage[utf8]{inputenc} 
\usepackage[T1]{fontenc}    
\usepackage{url}            
\usepackage{booktabs}       
\usepackage{amsfonts}       
\usepackage{nicefrac}       
\usepackage{microtype}      
\usepackage{xcolor}         
\usepackage{amsmath}
\usepackage[pagebackref=true,breaklinks=true,letterpaper=true,colorlinks,bookmarks=false]{hyperref}
\usepackage{graphicx, wrapfig}
\usepackage{booktabs}
\usepackage{array}
\usepackage{pgfplots}
\usetikzlibrary{pgfplots.groupplots}
\usepackage{array}
\usepackage{multirow}
\usepackage{placeins}
\usepackage{dsfont}
\usepackage{mathtools, stmaryrd}
\usepackage{caption}
\usepackage{subcaption}
\setcitestyle{numbers}
\setcitestyle{square}
\usepackage{amssymb}
\usepackage{svg}
\usepackage{xparse} \DeclarePairedDelimiterX{\Iintv}[1]{\llbracket}{\rrbracket}{\iintvargs{#1}}
\title{LLM meets Vision-Language Models for Zero-Shot One-Class Classification} 

\author{Yassir Bendou \\ IMT Atlantique, Brest, France \\ \texttt{yassir.bendou@imt-atlantique.fr} \And Giulia Lioi \\ IMT Atlantique, Brest, France \\ \texttt{giulia.lioi@imt-atlantique.fr} \And Bastien Pasdeloup \\ IMT Atlantique, Brest, France \\ \texttt{bastien.pasdeloup@imt-atlantique.fr} \And Lukas Mauch \\ Sony Europe, B.V. \\ Stuttgart Laboratory 1, Germany \\ \texttt{lukas.mauch@sony.com} \And Fabien Cardinaux \\ Sony Europe, B.V. \\ Stuttgart Laboratory 1, Germany \\ \texttt{fabien.cardinaux@sony.com} \And Ghouthi Boukli Hacene \\ Sony Europe, B.V. \\ Stuttgart Laboratory 1, Germany \\ Mila, Montreal, Canada\\ \texttt{ghouthi.bouklihacene@sony.com} \And Vincent Gripon \\ IMT Atlantique, Brest, France \\ \texttt{vincent.gripon@imt-atlantique.fr}}

\begin{document}
\maketitle
\begin{abstract}
We consider the problem of zero-shot one-class visual classification, extending traditional one-class classification to scenarios where only the label of the target class is available. This method aims to discriminate between positive and negative query samples without requiring examples from the target class. We propose a two-step solution that first queries large language models for visually confusing objects and then relies on vision-language pre-trained models (\emph{e.g.}, CLIP) to perform classification.
By adapting large-scale vision benchmarks, we demonstrate the ability of the proposed method to outperform adapted off-the-shelf alternatives in this setting.
Namely, we propose a realistic benchmark where negative query samples are drawn from the same original dataset as positive ones, including a granularity-controlled version of iNaturalist, where negative samples are at a fixed distance in the taxonomy tree from the positive ones.
To our knowledge, we are the first to demonstrate the ability to discriminate a single category from other semantically related ones using only its label. Our code is available online: \href{https://github.com/ybendou/one-class-ZS}{https://github.com/ybendou/one-class-ZS}
\end{abstract}
\section{Introduction}
\label{sec:intro}



One-class classification aims to obtain accurate discriminators for a single class~\cite{deeponeclass, oneclasssurvey}. This paradigm has been extensively studied as it encapsulates multiple research fields~\cite{oneclasssvm,oneclasssemi,explainableoneclass,han2022adbench}, including the well-known problem of anomaly detection where the aim is to learn a model that accurately identifies outliers. One-class classification finds its utility in various domains such as fraud detection~\cite{frauddetection}, healthcare~\cite{medicalanomaly}, cybersecurity~\cite{anomalycybersecurity} and damage detection~\cite{winclip}. Another practical application is wake-word detection in edge devices where the goal is to detect a certain signal that triggers some action from the system. 

Existing methods require gathering training samples to train a one-class classifier for a specific application~\cite{oneclasssurvey}. These methods are not flexible as they can not be easily adapted to new concepts when the class of interest changes without additional training samples~\cite{noveltydetection}. With the emergence of zero-shot classification using vision-language models such as CLIP~\cite{clip}, quick adaptation to new concepts has known a large interest in the multi-class case~\cite{clipseg,clip3d}. However, existing one-class classifiers are still bound to the traditional framework, struggling to rapidly adapt to the case where the target class could vary, requiring to retrain each time a new classifier.

In this work, we aim to extend one-class classification to include the open-vocabulary case, namely \emph{zero-shot one-class classification}, which allows for easy adaptation to different target classes. In this setting, only the target class is known and the objective is to accurately accept or reject data, as illustrated in Figure~\ref{fig:main}. This method is particularly valuable in industries where collecting training data for a ``special'' class is costly or in robotic systems that demand the flexibility to dynamically adjust the target class such as in real-time object detection~\cite{objectdetection}. Despite its potential, this area remains underexplored in the existing literature.

It is pertinent to contrast zero-shot one-class classification with classical image-text retrieval tasks, which operate under different assumptions. In classical image-text retrieval, large databases of query images are leveraged to retrieve the top-K queries based on a predefined similarity score~\cite{textimageretrieval1, textimageretrieval2}. Such a methodology relies on an extensive offline database, which contrasts with the online nature of zero-shot one-class classification. In our framework, each query is processed individually and is not part of an existing database, focusing solely on determining if a single image belongs to the target class. This capability is particularly advantageous in scenarios such as robotic devices, where the classification needs can change in real-time. 

Interestingly, existing zero-shot classification approaches based on CLIP~\cite{clip, multimodality, coop} are not easily adapted to this limit case. These methods require multiple categories to build their classifiers. Such a set of all negative categories at test time is often not available or not well defined. While classifying multiple objects typically involves estimating the boundaries between classes, one-class classification requires defining these boundaries solely based on examples from the class in question. Recent works also consider the case of zero-shot out of distribution~\cite{mcm,clipn,zoc}, mainly reporting AUC scores where the exact boundary estimation, through the use of an explicit threshold, is not directly addressed. However, we argue that validation samples are not necessarily available in practice, particularly in zero-shot classification, and that the question of finding the absolute position of the boundary still holds untapped potential.

\begin{figure}[t]
    \centering
    \subfloat[\centering Target class is ``Finch'']{\includegraphics[width=1.\textwidth]{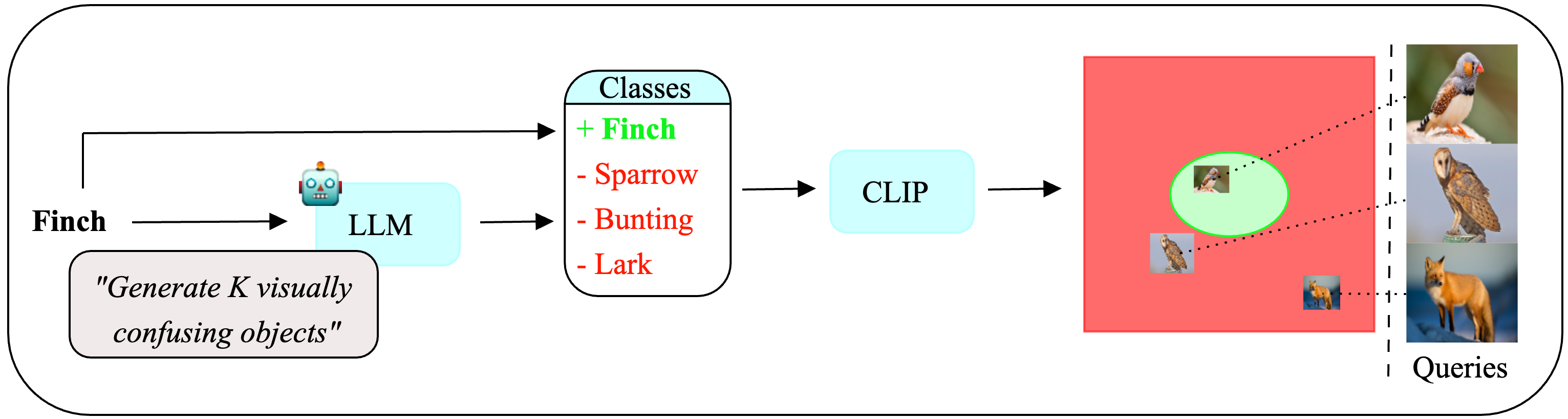}}\hfil 
    \subfloat[\centering Target class is ``Bird'']{\includegraphics[width=1.\textwidth]{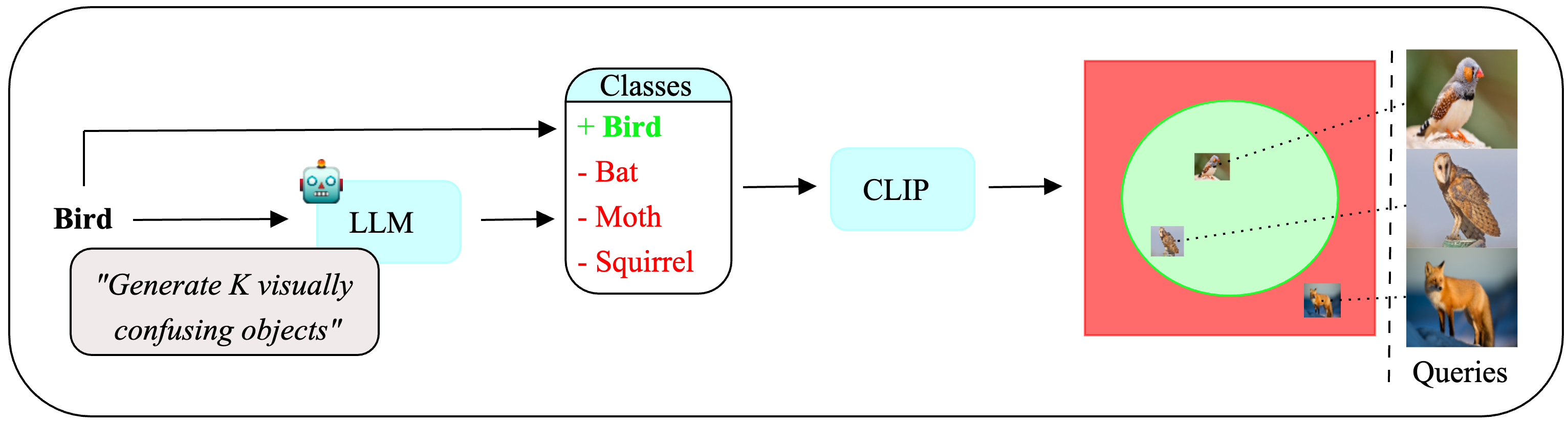}}\hfil 
    \caption{Illustration of the proposed method for two instances of a zero-shot one-class classification task, where only the class label is given. In the first case (a), only the ``Finch'' example is identified as a ``Finch''. In the second scenario (b), both bird examples are classified as birds. Negative examples of classes are obtained through querying an LLM. Notice that the position of query images is unchanged between the two instances, as only the class boundaries vary.}
    \label{fig:intro}
\end{figure}

Estimating the boundary of each class would require either collecting training samples or defining the set of all possible negative classes which in practice is intractable. We instead elaborate a simple, yet very effective solution, by collecting a set of high quality negative labels to define the boundary of the class. Such a set is constructed by prompting a Large Language Model (LLM) for the most visually confusing objects, as depicted in Figure~\ref{fig:intro}. This allows to construct a corpus of negative-labels adapted for target task. Using the similarity between text and images, we construct adaptive decision boundaries using the target class and the negative labels. We further regularize this decision boundary in order to restrain the positive samples around the target class. We run extensive experiments to validate our methodology against adapted off-the-shelf baselines and show its robustness on multiple settings including a granularity-controlled version of iNaturalist. Our findings show the importance of having adaptive decision boundaries based on each task rather that relying on a single threshold, especially on fine-grained and coarse tasks. Our contributions are as follows:

\begin{figure}[t!]
    \centering
        \includegraphics[width=1\textwidth]{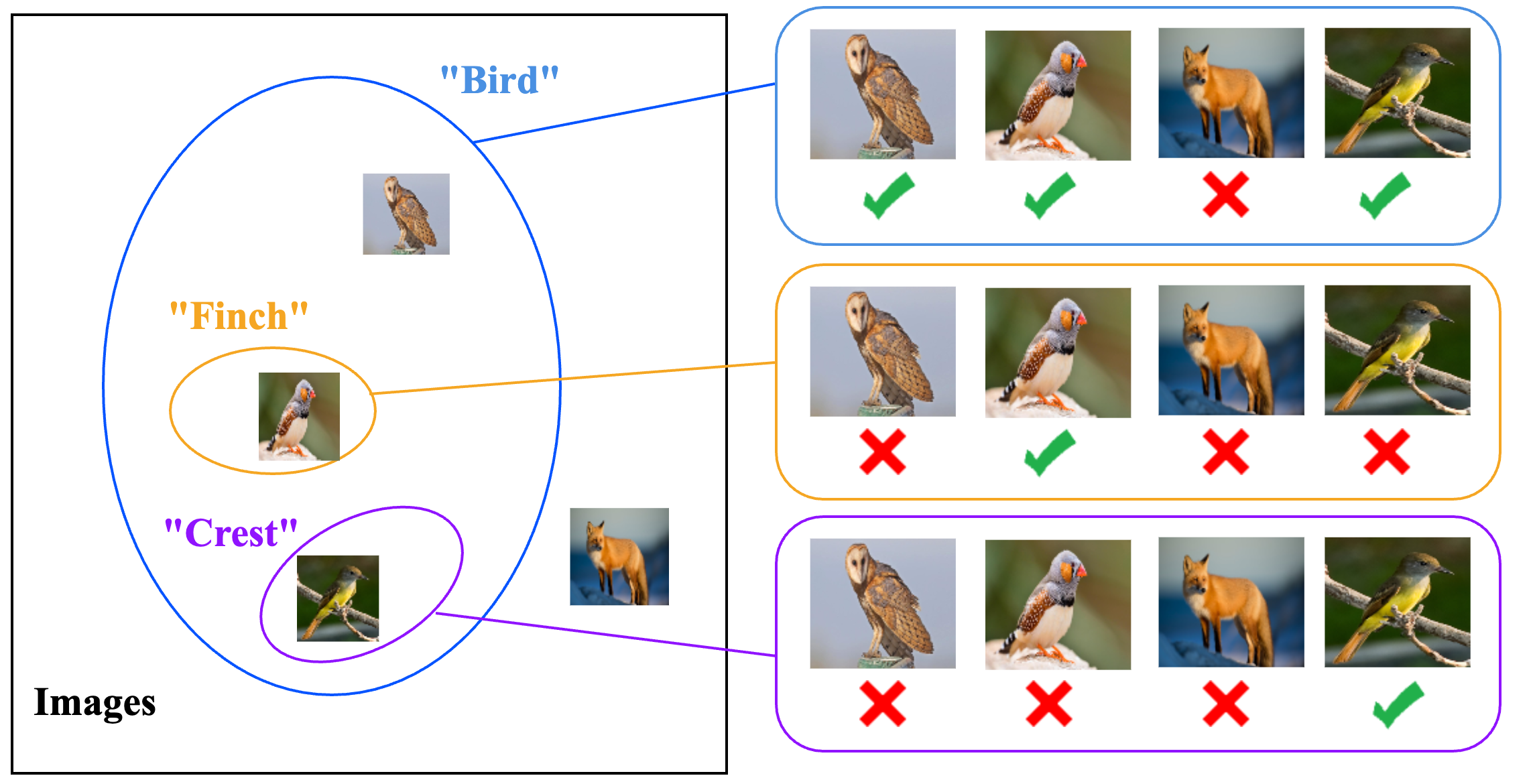}
    \caption{In the zero-shot one-class classification problem, the class is defined using a text prompt and the task is to accept or reject query images independently. As the prompt text varies, the decision boundaries of the class also varies. Such a task is challenging as it requires an accurate estimate of the boundaries of each class.}
    \label{fig:main}
\end{figure}

\begin{enumerate}
\item To the best of our knowledge, we are the first to tackle the problem of zero-shot one-class classification, where only the label of the class of interest is available;
\item We propose a realistic benchmark where negative query samples are drawn from the same original dataset as positive ones. We include a granularity-controlled version of iNaturalist, where negative samples are at a fixed distance in the taxonomy tree from the positive ones;
\item We propose a methodology making use of an LLM and a vision-language pre-trained model. We conduct extensive experiments to demonstrate the robustness of our approach across a diverse range of tasks, showing its benefits compared to adapted off-the-shelf alternatives.
\end{enumerate}

\section{Related Work}
\subsection{One-class Classification}
One-class classification is a binary classification task where only examples from the positive class are available~\cite{oneclasssurvey}. Deep one-class classification with neural networks was first introduced using an SVM for anomaly detection~\cite{deeponeclass} by mapping positive examples into a hypersphere and considering deviations from the center as anomalies. However, this method suffered mainly from representation collapse. An alternative approach tackles this issue by generating negative samples in a lower dimensional manifold~\cite{drocc}. Other semi-supervised methods have been also proposed, but they often have access to few negative samples~\cite{oneclasssemi,semisup2,semisup3}. GANs have also been used for one-class classification~\cite{gan1,gan2,gan3}. In a similar line of work, the use auto-encoders has been also explored by training using only the positive samples~\cite{autoencoder1,autoencoder2,autoencoder3}. In our work, we investigate zero-shot scenarios where no training samples are available except for the target class label. 

\subsection{Large-scale Out-of-Distribution (OOD) Detection}
The closest literature to our work is OOD detection. In this setting, OOD data belong to unknown classes while positive samples belong to a set of closed classes. This allows to extend the capabilities of multi-class classifiers to detect instances from unknown classes~\cite{openset,calibrate2}. The goal of these methods is to propose a confidence score to reject the OOD data~\cite{softmax,odin,enerybasedOOD,scaling,maxlogit,negativeprototype,opensetboudiaf,locoop, mcm}. However, most of previously mentioned methods assume the existence of a validation set and do not tackle the challenge of selecting an appropriate threshold. This question is often disregarded by using evaluation metrics that are threshold-invariant (\emph{e.g.}, AUC score) or depend on a validation set (\emph{e.g.}, False Positive Rate at 95\% True Positive Rate)~\cite{calibrate3,calibrate4}. Threshold-free methods have been proposed to train a classifier to learn adaptive thresholds in~\cite{placeholders}, however it requires having access to in-domain (ID) data~\cite{negativeprototype}. To our knowledge, CLIPN~\cite{clipn} and ZOC~\cite{zoc} are the only zero-shot out-of-distribution methods which do not require validation samples and can be easily adapted to zero-shot one-class classification. The authors of CLIPN fine-tune CLIP to generate negative prompts but only test their method on ``easy'' tasks where ID samples are drawn from a dataset and OOD samples are drawn from another one. Similarly, ZOC employs BERT to predict the names of the OOD classes and estimates an anomaly score based on the cosine similarity to the negative classes. Unlike CLIPN and ZOC which primarily reports threshold-invariant metrics such as AUC score, our work undertakes a direct comparison of the acceptance/rejection performance with these two baselines.

\section{Preliminaries}
\label{sec:preliminaries}

\subsection*{Problem Statement}

Let us formally define the zero-shot one-class classification task: the goal is to learn a classification function $\psi_{\mathbf{t}}(\mathbf{x})$ that maps an input $\mathbf{x}$ to its label $y\in\{0,1\}$ using only the text $\mathbf{t}$ of the target class. 

Our objective is to build a binary classifier that aims at maximizing the macro F1 score, defined as $\text{F1}_{\text{macro}} = \frac{\text{F1}_{+} + \text{F1}_{-}}{2}$, where $\text{F1}_{+}$ and $\text{F1}_{-}$ are the F1 scores of the positive and negative classes respectively. This is a classical measure of performance in one-class classification and OOD detection~\cite{negativeprototype}.

We expect positive and negative samples to be drawn from very different underlying distributions, since positive samples typically represent a single category, whereas negatives can be drawn from multiple categories. This asymmetry is important, as it suggests the use of classifiers where positive and negative class domains can be very different.

\subsection*{Vision-Language Pre-Training}
In recent years, visual language models have gained traction, integrating text and images for more effective learning. Contemporary models like CLIP~\cite{clip} and ALIGN~\cite{align} use a contrastive learning framework, showing impressive multi-class zero-shot learning capabilities without additional fine-tuning. CLIP demonstrates a straightforward yet powerful approach by training a double encoder network $\mathcal{T}\colon \mathbf{t}\mapsto \mathbf{w} \in \mathbb{R}^d$ and $\mathcal{I}\colon \mathbf{x}\mapsto \mathbf{v} \in \mathbb{R}^d$ to embed text and images respectively, in a shared embedded space. By treating the class label as a class prototype, this approach allows to compute a cosine similarity score between image and text embeddings for zero-shot classification defined as: $s(\mathbf{x},\mathbf{t})=\frac{\mathcal{I}(\mathbf{x})^{\top}\mathcal{T}(\mathbf{t})}{\|\mathcal{I}(\mathbf{x})\|\|\mathcal{T}(\mathbf{t})\|}$. In the next section, we discuss how to build a one-class classifier based on this similarity.

\section{Methodology}

Leveraging the alignment between text and images in CLIP-like architectures~\cite{clip}, we build a one-class classifier using the cosine similarity between the target class~$\mathbf{t}$ and the query image $\mathbf{x}$. Formally, given a threshold $\lambda_{\mathbf{t}}(\mathbf{x})$, the zero-shot one-class classification is performed as: 

\begin{align}
\psi_{\mathbf{t}}(\mathbf{x}) = \left\{
                \begin{array}{ll}
                  1 \quad \text{if} \quad s(\mathbf{x},\mathbf{t}) \geq \lambda_{\mathbf{t}}(\mathbf{x}),  \\
                  0 \quad \text{otherwise} 
                \end{array}
              \right.
\end{align}
This formulation is similar to the score function used in OOD detection~\cite{mcm,clipn}. The main difference is that in OOD detection, the choice of the threshold is not discussed as pointed by~\cite{negativeprototype}. In the case of one-class classification, we are interested at setting an appropriate threshold for each task. 

To this end, we devise two possible choices for $\lambda_{\mathbf{t}}(\mathbf{x})$. A first simple yet effective strategy is to select a constant threshold for all tasks. However, such a strategy assumes that all tasks have the same threshold level. The second approach is to design an adaptive threshold that depends on each task and allows to effectively capture the varied nature of the tasks.

\subsection*{Fixed Thresholding}
The first method we consider consists in using a single threshold for all tasks, leading to a constant decision boundary. Given that we do not have a validation set for the downstream task, we propose to use ImageNet1K to compute an average threshold from a large number of artificially generated tasks. Note that this dataset is often used for selecting hyperparameters in multi-class zero-shot classification~\cite{imagenettransfer1,imagenettransfer2}. The threshold is selected as the one which maximizes the F1 score over $1000$ one-class tasks. We refer to this first method as Fixed Thresholding (FT).

\begin{wrapfigure}{r}{0.5\textwidth}
    \centering
    \resizebox{0.5\textwidth}{!}{{\input{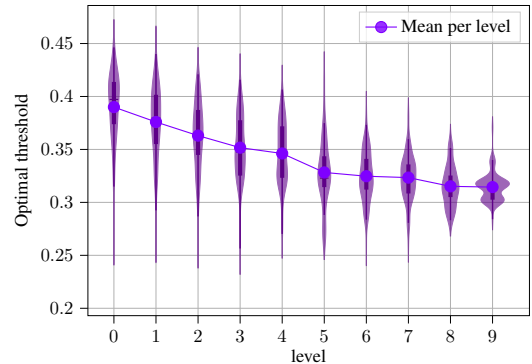}}}
    \captionof{figure}{Distribution of the optimal threshold for each task for different levels in iNaturalist.}
    \label{fig:optimal_threshold}
    \vspace{-1cm}
\end{wrapfigure}

Our experiments show that, while fixing a single threshold $\bar{\lambda}$ for all tasks is a reasonable baseline in some cases, an optimal threshold for a certain task may fail for other tasks, especially for scenarios that are far from the average ImageNet1K ranges. In fact, different classes can have different ranges of cosine similarities between the target class and its corresponding images. We showcase this effect in Figure~\ref{fig:optimal_threshold} where we plot the distribution of the optimal thresholds for various tasks with different granularities, motivating the need for an adaptive threshold method.

\subsection*{Adaptive Thresholding}

In order to obtain adaptive boundary decisions, a recent method proposed in the few-shot setting to use negative prototypes~\cite{negativeprototype}. The aim is to obtain a set of negative prototypes to serve as rejection boundaries for different tasks. The query image is then rejected if the similarity to the target class is lower than that of the negative prototypes. Since generating these negative classes from the training data is infeasible in the zero-shot setting, we instead use an LLM to generate them. To do so, we prompt GPT-4~\cite{gpt4} to obtain $K$ labels that could be visually confused with the target class. We refer to the $K$ visually confusing objects $(\mathbf{t}_{n_1}, \dots, \mathbf{t}_{n_K})$. We then propose two methods to exploit these negative labels.

The first one is to compute $K$ thresholds and take their maximum, which we call Multi Negative Prototype (MNP): 

\begin{align}
    \lambda_{\mathbf{t}}(\mathbf{x})~=~\max_{k \in \{1 \dots K\}} \frac{\mathcal{I}(\mathbf{x})^{\top}\mathcal{T}(\mathbf{t}_{n_k})}{\|\mathcal{I}(\mathbf{x})\|\|\mathcal{T}(\mathbf{t}_{n_k})\|}
\end{align}
In essence, the query is rejected if it matches any of the negative labels. Although this method allows to dynamically estimate the rejection boundary, the risk of false negatives may occur. In fact, the LLM might output adversarial labels that could match more than the positive. To avoid this, we smooth the predictions of the LLM by averaging the negative prototypes in the feature space which allows to get a more robust estimate of the negative prototype:
\begin{align}
\label{eq:ANT}
    \lambda_{\mathbf{t}}(\mathbf{x}) = \frac{\mathcal{I}(\mathbf{x})^{\top}\bar{\mathbf{w}}}{\|\mathcal{I}(\mathbf{x})\|\|\bar{\mathbf{w}}\|} \quad \text{where} \quad \bar{\mathbf{w}} ~=~\sum_{i=1}^{K}{\frac{\mathcal{T}\left(\mathbf{t}_{n_i}\right)}{K}}
\end{align}
As shown in Section~\ref{sec:ablations}, this method which we refer to as Average Negative Prototype (ANP) is more robust than MNP.

\subsection*{Adaptive $+$ Fixed Thresholding}
So far, we have shed light on constructing adaptive decision boundaries using negative labels. Despite being able to adapt to different ranges of tasks, using negative labels has a limitation. Essentially, if a sample is distant from the negative prototypes, it is deemed positive regardless of its closeness to the positive class prototype. This characteristic can lead to a broader, less precise definition of the positive class, potentially including false positives. Such limitation raises the necessity to restrain the positive predictions within a specific region relative to the positive prototype.

To do so, we propose to combine adaptive and fixed thresholding methods. This hybrid approach aims to refine the classification by incorporating the fixed threshold obtained from ImageNet1K~(\emph{i.e.}, $\bar{\lambda}$) alongside the adaptive thresholding:
\begin{align}
\label{eq:combined_threshold}
    \bar{\lambda}_{\mathbf{t}}(\mathbf{x})=\alpha\lambda_{\mathbf{t}}(\mathbf{x})+(1-\alpha)\bar{\lambda} \quad \text{where} \quad \alpha \in [0,1]
\end{align} 
This combined threshold serves to more precisely delineate the positive class, restricting it to a more defined region. By doing so, this method ensures that samples are classified as positive not only by being distant from the negative prototypes but also by being genuinely close to the target prototype. Considering the typically positive nature of cosine similarities between text and image representations in CLIP~\cite{positiveclip}, natural images $\mathbf{x}$ satisfy the following: $\bar{\lambda}_{\mathbf{t}}(\mathbf{x}) \geq (1-\alpha)\bar{\lambda}$. Accordingly, the positive classification is effectively bounded within a region exhibiting a higher similarity than $(1-\alpha)\bar{\lambda}$, ensuring a more precise and restrictive definition of the positive class. For a hyperparamter-free method, we set $\alpha$ to be $\frac{1}{2}$. The choice of $\alpha$ is further discussed in the experiments section.


\section{Proposed Benchmark}
\label{sec:sampling}

A simple way to generate realistic one-class classification benchmarks consists in using existing classification vision datasets. We randomly select a class $c$ to be our target, and sample positive and negative examples from respectively the target class and the other ones with a ratio of positive samples~$r$. We refer to this sampling as ``uniform'' sampling. However, by doing so we do not have explicit control over the granularity of the generated tasks which is important for a balanced evaluation. We thus also propose a ``hierarchical'' sampling based on the taxonomy of the iNaturalist~\cite{inaturalist} dataset allowing to control the proximity between positive and negative examples. 

The hierarchy of iNaturalist forms a tree that begins with the most inclusive categories (\emph{i.e.}, ``Kingdoms'') which encompass broad groups such as plants, animals, and fungi. From there, it progressively narrows down through various levels of specificity until the individual species that constitute the ten thousands of leaves of the tree.

Considering a level of granularity $\ell$, we generate tasks by first selecting a leaf uniformly at random in the tree. Its $\ell$-th ancestor becomes the class of positive examples, and its $\ell+1$-th ancestor, deprived of the $\ell$-th ancestor, becomes the class of negative examples. The motivation behind this choice is to avoid overly simplistic tasks such as discriminating between a cell and a dog. The proposed sampling creates hard tasks that we believe to be realistic. Finally, for each task, we sample 100 queries with a fixed ratio of positive samples $r$.

\section{Experimental Setup}
\subsection{Datasets and Metrics}
In our experiments, we employ the hierarchical sampling for iNaturalist and the uniform sampling for 5 existing datasets: EuroSAT~\cite{eurosat}, Food~\cite{food101}, SUN~\cite{sun397}, Textures~\cite{dtd} and Pets\cite{oxford_pets}. We sample 100 queries per task and report the performance over 1,000 runs. The ratio of positive samples is set to $r=0.5$, unless specified otherwise. We primarily report our results using the macro F1 score, a classical measure in the field of out-of-distribution detection. To provide a broader picture of our methods' effectiveness, we also report the AUC score in the supplementary material as well as the accuracy for cases with balanced data.

\subsection{Implementation Details and Baselines}
For a fair comparison with adapted off-the-shelf alternatives, we use the same architecture -- \emph{i.e.}, the ViT-B-32 from CLIP~\cite{clip} -- which is commonly used for zero-shot tasks. We use use the template ``a photo of a <class>'' to extract the textual features. In addition, the use of scientific names in the iNaturalist dataset has been demonstrated to yield suboptimal results. This is because models like CLIP were predominantly trained on common names~\cite{scientificnamesinaturalist}. Similarly to previous work~\cite{scientificnamesinaturalist}, to address this, we opt for common names obtained using GPT4 in our experiments. We include in the supplementary material a comparison between common and scientific names. These names are standardized across all compared methods for fairness. We query the considered LLM for generating 10 negative prototypes which we vary in the supplementary material. All our experiments are performed using an NVIDIA RTX 3090. 

We compare our method to two existing alternatives from zero-shot OOD detection which we adapt to the zero-shot one-class case. The first method is CLIPN, a variant of CLIP enhanced to generate negative prompts using the template ``a photo of no <class>''~\cite{clipn}. The second approach is ZOC~\cite{zoc}, a fine-tuned BERT decoder on COCO dataset~\cite{coco} to generate class candidates potentially present in the query image. These two methods generate negative prototypes which could be seen as an alternative to the adaptive threshold module. Therefore, we propose to combine them with the fixed thresholding regularization resulting in CLIPN+FT and ZOC+FT.

\section{Results}
\label{sec:results}
\subsection{Performance}
We compare our proposed method (ANP+FT) with other alternatives in Table~\ref{tab:main_performance}. We observe that our proposed method achieves state-of-the-art performance across all of the considered datasets and backbones, with the noticeable exception of EuroSAT -- which can be explained by its low-resolution images -- resulting in a significant gain in average compared to considered baselines. It is noteworthy to add that ZOC requires significantly more computations at inference time since it uses an additional BERT decoder for each query image. On the hand our method significantly outperforms CLIPN which uses a finetuned version of CLIP. 





\begin{table}[htbp!]
  \caption{Average macro F1 score for multiple datasets.
  }
   \label{tab:main_performance}
  \centering
  \resizebox{\linewidth}{!}{\begin{tabular}{@{}c|cccccc|c@{}}
    \toprule 
Method &	iNat	& SUN	& Food	& Textures	& Pets & EuroSAT	& Average \\
\midrule

CLIPN~\cite{clipn}	&$42.53\pm0.73$&$82.14\pm0.55$&$68.77\pm0.80$&$52.39\pm0.66$&$66.01\pm0.82$&$64.58\pm1.08$&$62.74$\\

ZOC\cite{zoc} &$55.05\pm0.99$&$93.61\pm0.32$&$92.18\pm0.24$&$73.88\pm0.66$&$88.53\pm0.77$&$\mathbf{65.38\pm1.14}$&$78.16$\\
{ANP+FT} (ours) &$\mathbf{60.32\pm0.99}$&$\mathbf{94.19\pm0.38}$&$\mathbf{94.18\pm0.20}$&$\mathbf{78.21\pm0.88}$&$\mathbf{90.97\pm0.88}$&$59.72\pm0.98$&$\mathbf{79.60}$\\
\bottomrule
\end{tabular}}
\end{table}

\begin{figure}[htbp!]
    \centering
    \begin{minipage}{0.48\textwidth}
        \resizebox{1\textwidth}{!}{\begin{tikzpicture}

\definecolor{darkgray176}{RGB}{176,176,176}
\definecolor{green}{RGB}{0,128,0}
\definecolor{lightgray204}{RGB}{204,204,204}
\definecolor{pink}{RGB}{255,192,203}

\begin{axis}[
legend cell align={left},
legend style={
  fill opacity=0.8,
  draw opacity=1,
  text opacity=1,
  at={(0.03,0.97)},
  anchor=north west,
  draw=lightgray204
},
tick align=outside,
tick pos=left,
x grid style={darkgray176, },
xlabel={Level},
xmin=-0.45, xmax=9.45,
xtick style={color=black},
y grid style={darkgray176},
ylabel={macro F1 (\%)},
ymin=33, ymax=85,
ytick style={color=black},
ymajorgrids,
xmajorgrids,
xtick={0,1,2,3,4,5,6,7,8,9},
]
\path [draw=green, fill=green, opacity=0.5]
(axis cs:0,36.2414600098421)
--(axis cs:0,35.4631977928465)
--(axis cs:1,37.1916845204137)
--(axis cs:2,39.0138009293096)
--(axis cs:3,40.819578688927)
--(axis cs:4,41.1405262177671)
--(axis cs:5,38.1284473097081)
--(axis cs:6,41.464247427444)
--(axis cs:7,44.003473154809)
--(axis cs:8,47.8732530743134)
--(axis cs:9,52.9663154206593)
--(axis cs:9,54.8127593055688)
--(axis cs:9,54.8127593055688)
--(axis cs:8,49.8659381636414)
--(axis cs:7,45.6835325851511)
--(axis cs:6,43.0105624544208)
--(axis cs:5,39.3395789757214)
--(axis cs:4,42.7337120755166)
--(axis cs:3,42.3490247224686)
--(axis cs:2,40.3360835739537)
--(axis cs:1,38.2545408183953)
--(axis cs:0,36.2414600098421)
--cycle;


\addplot [semithick, green, mark=*, mark size=3, mark options={solid}]
table {%
0 35.8523289013443
1 37.7231126694045
2 39.6749422516317
3 41.5843017056978
4 41.9371191466419
5 38.7340131427147
6 42.2374049409324
7 44.8435028699801
8 48.8695956189774
9 53.8895373631141
};
\addlegendentry{CLIPN}
\path [draw=red, fill=red, opacity=0.5]
(axis cs:0,42.5540047500255)
--(axis cs:0,41.1835115505444)
--(axis cs:1,46.6053476932937)
--(axis cs:2,50.3713751477244)
--(axis cs:3,51.6557214948789)
--(axis cs:4,54.6074447471335)
--(axis cs:5,55.3649456967986)
--(axis cs:6,57.2744137451228)
--(axis cs:7,57.5128755655224)
--(axis cs:8,61.1461261058671)
--(axis cs:9,64.8596558650252)
--(axis cs:9,67.0352781256706)
--(axis cs:9,67.0352781256706)
--(axis cs:8,63.1635948761337)
--(axis cs:7,59.7101310848086)
--(axis cs:6,59.1548975574304)
--(axis cs:5,57.3733185744971)
--(axis cs:4,57.0217106571728)
--(axis cs:3,53.854249300223)
--(axis cs:2,52.2698334986155)
--(axis cs:1,48.2067629133509)
--(axis cs:0,42.5540047500255)
--cycle;


\path [draw=blue, fill=blue, opacity=0.5]
(axis cs:0,47.5120025623836)
--(axis cs:0,45.7907613599625)
--(axis cs:1,48.1311658975023)
--(axis cs:2,49.0021873783862)
--(axis cs:3,52.9713417302179)
--(axis cs:4,59.147239103802)
--(axis cs:5,58.9585786401696)
--(axis cs:6,61.565532921208)
--(axis cs:7,68.6689589990613)
--(axis cs:8,70.4234062958137)
--(axis cs:9,78.6261461255832)
--(axis cs:9,80.3401586334941)
--(axis cs:9,80.3401586334941)
--(axis cs:8,72.344995692739)
--(axis cs:7,70.6769975458387)
--(axis cs:6,63.68108794958)
--(axis cs:5,60.9415041737912)
--(axis cs:4,61.591657273127)
--(axis cs:3,55.1833706343627)
--(axis cs:2,50.8868988212431)
--(axis cs:1,49.9214801152445)
--(axis cs:0,47.5120025623836)
--cycle;

\addplot [semithick, red, mark=*, mark size=3, mark options={solid}]
table {%
0 41.868758150285
1 47.4060553033223
2 51.32060432317
3 52.7549853975509
4 55.8145777021531
5 56.3691321356478
6 58.2146556512766
7 58.6115033251655
8 62.1548604910004
9 65.9474669953479
};
\addlegendentry{ZOC}

\addplot [semithick, blue, mark=*, mark size=3, mark options={solid}]
table {%
0 46.6513819611731
1 49.0263230063734
2 49.9445430998146
3 54.0773561822903
4 60.3694481884645
5 59.9500414069804
6 62.623310435394
7 69.67297827245
8 71.3842009942764
9 79.4831523795387
};
\addlegendentry{ANP+FT (ours)}

\end{axis}

\end{tikzpicture}}
        \captionof{figure}{Macro F1 score per level in iNaturalist for different methods.}
        \label{fig:f1_per_level}
    \end{minipage}
    \hfill
    \begin{minipage}{0.48\textwidth}
        \centering
        \resizebox{1\textwidth}{!}{
\begin{tikzpicture}

\definecolor{darkgray176}{RGB}{176,176,176}
\definecolor{green}{RGB}{0,128,0}
\definecolor{lightgray204}{RGB}{204,204,204}
\definecolor{pink}{RGB}{255,192,203}

\begin{axis}[
width=10cm,
height=8cm,
legend cell align={left},
legend style={fill opacity=0.8, draw opacity=1, text opacity=1, draw=lightgray204, at={(0.97,0.03)},
  anchor=south east},
tick align=outside,
tick pos=left,
x grid style={darkgray176},
xlabel={Positive Rate (\%)},
xmajorgrids,
xmin=0, xmax=100,
xtick style={color=black},
y grid style={darkgray176},
ylabel={macro F1 (\%)},
ymajorgrids,
ymin=20, ymax=90,
ytick style={color=black},
xtick={0,10,20,30,40,50,60,70,80,90,100},
at={(0.97,0.03)},
]
\path [draw=green, fill=green, opacity=0.5]
(axis cs:10,40.1484963152456)
--(axis cs:10,38.5044075574644)
--(axis cs:50,61.9621618335017)
--(axis cs:90,70.2499976642277)
--(axis cs:90,71.7271162690963)
--(axis cs:90,71.7271162690963)
--(axis cs:50,63.5100906362007)
--(axis cs:10,40.1484963152456)
--cycle;


\addplot [semithick, green, mark=*, mark size=3, mark options={solid}]
table {%
10 39.326451936355
50 62.7361262348512
90 70.988556966662
};
\addlegendentry{CLIPN}
\path [fill=red, fill opacity=0.5]
(axis cs:90,75.004508972168)
--(axis cs:90,73.2844924926758)
--(axis cs:50,77.3700866699219)
--(axis cs:10,64.992790222168)
--(axis cs:10,66.4351272583008)
--(axis cs:10,66.4351272583008)
--(axis cs:50,78.7416839599609)
--(axis cs:90,75.004508972168)
--cycle;

\addplot [semithick, red, mark=*, mark size=3, mark options={solid}]
table {%
90 74.1445007324219
50 78.0558853149414
10 65.7139587402344
};
\addlegendentry{ZOC}
\path [draw=blue, fill=blue, opacity=0.5]
(axis cs:10,72.643814874804)
--(axis cs:10,70.9898793746423)
--(axis cs:50,78.8798634130995)
--(axis cs:90,70.2752095076496)
--(axis cs:90,72.2001597768626)
--(axis cs:90,72.2001597768626)
--(axis cs:50,80.3174287175828)
--(axis cs:10,72.643814874804)
--cycle;

\addplot [semithick, blue, mark=*, mark size=3, mark options={solid}]
table {%
10 71.8168471247232
50 79.5986460653412
90 71.2376846422561
};
\addlegendentry{ANP+FT}

\end{axis}

\end{tikzpicture}}
        \captionof{figure}{Macro F1 score for different positive rates.}
        \label{fig:positive_rate}
    \end{minipage}
\end{figure}

In order to better understand the effect of varying the task granularity, we report in Figure~\ref{fig:f1_per_level} the performance on iNaturalist per level. Note that a lower level means a fine-grained classification between very similar species, while higher levels represent larger and coarser concepts. Our results show that our method outperforms both CLIPN and ZOC. While ZOC is competitive on some levels (2 and 3), our method significantly outperforms it on other levels with higher or lower granularity. Generating negative prototypes using an LLM allows to effectively capture both fine-grained and coarse concepts. On the other hand, negative prototypes generated using ZOC are limited to the corpus of COCO on which it was fine-tuned, which can be particularly difficult when dealing with specific concepts (\emph{e.g.}, discriminating between ``Mugwort'' and ``Sagebrush''). We provide in the supplementary material some visual examples showcasing the difficulty of our proposed setting. On the other hand, our method is flexible and can be adapted to a large range of tasks.

\subsection{Results on imbalanced settings}

So far we have evaluated our method using a ratio of positive samples at $r=0.5$. We further assess the robustness of our method when this rate varies. Figure~\ref{fig:positive_rate} shows that our method is robust to data imbalance compared to the alternatives. CLIPN~\cite{clipn} fails in very small positive rate regimes, which highlights its sensitivity to false positives. On the other hand, while ZOC performs better with more false negatives, it also struggles with false positives similarly to CLIPN. Overall, our method maintains a better balance between false positives and false negatives.

\subsection{Effect of regularization}
Next, we report the performance of CLIPN and ZOC when combined with FT in Table~\ref{tab:boosts}. We observe a significant improvement of $13.95\%$ for CLIPN. On the other hand, ZOC benefits less from the regularization with varying gains depending on each dataset. While CLIPN and our method both generate negative prototypes from the target class, ZOC generates the negative prototypes from the queries. This can make it less sensitive to false positives, limiting the necessity of constraining positive predictions in a high cosine similarity region. On average, our method surpasses CLIPN+FT by $2.91\%$ which demonstrates the effectiveness of using the negative labels from the LLM instead of generating them from the finetuned version of CLIP.



\begin{table}[htbp!]
  \caption{Effect of regularization for existing baselines. Average macro F1 score.
  }
   \label{tab:boosts}
  \centering
  \resizebox{\linewidth}{!}{
  \begin{tabular}{@{}c|cccccc|c@{}}
    \toprule 
Method &	iNat	& SUN	& Food	& Textures	& Pets & EuroSAT	& Average \\
\midrule
CLIPN\cite{clipn}	&$42.53\pm0.73$&$82.14\pm0.55$&$68.77\pm0.80$&$52.39\pm0.66$&$66.01\pm0.82$&$64.58\pm1.08$&$62.74$\\
CLIPN\cite{clipn} + {FT}	
&$50.03\pm1.02$&$94.69\pm0.22$&$87.52\pm0.40$&$72.08\pm0.78$&$82.22\pm0.72$&$73.60\pm0.76$&$76.69$\\
ZOC\cite{zoc}	&$55.05\pm0.99$&$93.61\pm0.32$&$92.18\pm0.24$&$73.88\pm0.66$&$88.53\pm0.77$&$\mathbf{65.38\pm1.14}$&$78.16$\\
ZOC\cite{zoc}+{FT}	&$53.78\pm0.10$&$\mathbf{95.30\pm0.28}$&$93.74\pm0.21$&${77.51\pm0.68}$&$87.91\pm0.76$&$64.82\pm1.05$&$78.43$\\
{ANP+FT} &$\mathbf{60.32\pm0.99}$&$94.19\pm0.38$&$\mathbf{94.18\pm0.20}$&$\mathbf{78.21\pm0.88}$&$\mathbf{90.97\pm0.88}$&$59.72\pm0.98$&$\mathbf{79.60}$\\
\bottomrule
\end{tabular}}
\end{table}

\subsection{How well does a fixed threshold transfer from ImageNet1K?}
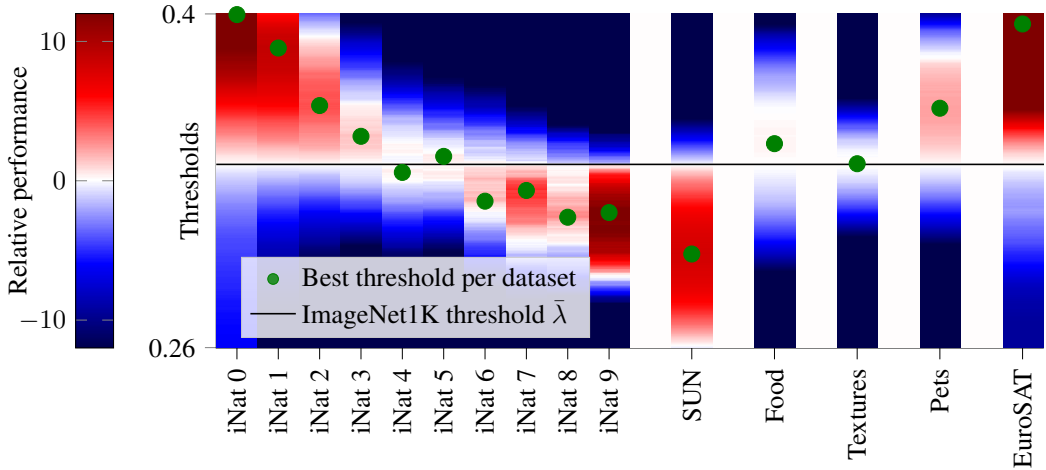
\begin{figure}[htbp!]
    \centering
    \resizebox{\textwidth}{!}{{
\begin{tikzpicture}

\definecolor{darkgray176}{RGB}{176,176,176}
\definecolor{green}{RGB}{0,128,0}
\definecolor{lightgray204}{RGB}{204,204,204}

\begin{axis}[
width=12.5cm,
height=6cm,
colorbar,
colorbar left,
colorbar style={ylabel={Relative performance},xshift=-0.05cm, ylabel style={at={(1,0.5)}},}, 
colormap={mymap}{[1pt]
  rgb(0pt)=(0,0,0.3);
  rgb(1pt)=(0,0,1);
  rgb(2pt)=(1,1,1);
  rgb(3pt)=(1,0,0);
  rgb(4pt)=(0.5,0,0)
},
legend cell align={left},
legend style={
  fill opacity=0.8,
  draw opacity=1,
  text opacity=1,
  at={(0.03,0.03)},
  anchor=south west,
  draw=lightgray204
},
point meta max=12,
point meta min=-12,
tick align=outside,
tick pos=left,
x grid style={darkgray176},
xmin=-0.5, xmax=19.5,
xtick style={color=black},
xtick={0,1,2,3,4,5,6,7,8,9,11,13,15,17,19},
xticklabel style={rotate=90.0},
xticklabels={
  iNat 0,
  iNat 1,
  iNat 2,
  iNat 3,
  iNat 4,
  iNat 5,
  iNat 6,
  iNat 7,
  iNat 8,
  iNat 9,
  SUN,
  Food,
  Textures,
  Pets,
  EuroSAT 
},
y dir=reverse,
ylabel={Thresholds},
y label style={at={(0.08,0.5)}},
ymin=-0.5, ymax=499.5,
ytick={0,499,500},
yticklabels={0.4,0.26,0.26}
]
\addplot [draw=green, fill=green, mark=*, only marks]
table{%
x  y
0 1
1 51
2 137
3 183
4 237
5 213
6 280
7 264
8 304
9 297
10 nan
11 359
12 nan
13 194
14 nan
15 224
16 nan
17 141
18 nan
19 15
};
\addlegendentry{Best threshold per dataset}


\addplot [semithick, draw=black]
table {%
-0.5 225
20.5 225
};
\addlegendentry{ImageNet1K threshold $\bar{\lambda}$}

\addplot graphics [includegraphics cmd=\pgfimage,xmin=-0.5, xmax=19.5, ymin=499.5, ymax=-0.5] {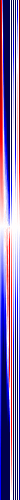};
\addplot [draw=green, fill=green, mark=*, only marks, forget plot, mark size=3pt]
table{%
x  y
0 1
1 51
2 137
3 183
4 237
5 213
6 280
7 264
8 304
9 297
11 359
13 194
15 224
17 141
19 15
};

\addplot [semithick, draw=black, forget plot]
table {%
-0.5 225
20.5 225
};

\end{axis}
\end{tikzpicture}}}
    \caption{Impact of varying the fixed threshold, for each dataset and for each level for iNaturalist. We report the relative macro F1 score to the one obtained using the ImageNet1K threshold $\bar{\lambda}$.}
    \label{fig:heatmap}
\end{figure}
In order to investigate the transferability of a fixed threshold from ImageNet1K to other datasets, we vary in Figure~\ref{fig:heatmap} the selected threshold and report its relative macro F1 score to the FT method. We observe that a single threshold is not optimal across different datasets. For iNaturalist, the wider the concept is, the lower are the similarities between images and the class labels. In contrast, fine-grained tasks tend to have high similarities with their class labels. Interestingly, some datasets/levels are more sensitive to the threshold choice than others. For example, level 9 of iNaturalist and Textures have a small range of well performing values while EuroSAT and SUN have larger ranges. Furthermore, although the threshold transferred from ImageNet1K is a good fit for some intermediate levels of iNaturalist and Food and Textures, it remains far from the optimal threshold for the others ones.


\section{Ablations}
\label{sec:ablations}
\begin{figure}[htbp!]
    \centering
    \begin{minipage}{0.48\textwidth}
        \resizebox{0.9\textwidth}{!}{\begin{tabular}{@{}c|c|c|c@{}}
  \toprule
Method & \parbox{1.65cm}{\centering Negative prototype} & \parbox{1.65cm}{\centering  Fixed threshold} & F1 score \\
\midrule
FT&&\checkmark&$78.81$\\
MNP&\checkmark&&$65.56$\\
ANP&\checkmark&&$66.23$\\
MNP+FT&\checkmark&\checkmark&$78.73$\\
ANP+FT&\checkmark&\checkmark&$\mathbf{79.60}$\\
\bottomrule
\end{tabular}



 


 

}
        \captionof{table}{Ablation study showing the average macro F1 scores for different modules.}
        \label{tab:ablation_table}
    \end{minipage}
    \hfill
    \begin{minipage}{0.48\textwidth}
       \centering
        \resizebox{0.9\textwidth}{!}{
\begin{tikzpicture}

\definecolor{darkgray176}{RGB}{176,176,176}
\definecolor{steelblue31119180}{RGB}{31,119,180}
\definecolor{green}{RGB}{0,128,0}

\begin{axis}[
height=8cm, 
width=8cm,
tick align=outside,
tick pos=left,
x grid style={darkgray176},
xlabel={Interpolation coefficient ($\alpha$)},
xmin=-0.05, xmax=1.05,
xtick style={color=black},
y grid style={darkgray176},
ylabel={macro F1 (\%)},
ymin=50, ymax=85,
ytick style={color=black},
legend style={
  fill opacity=0.8,
  draw opacity=1,
  text opacity=1,
  at={(0.03,0.1)},
  anchor=south west
},
ymajorgrids,
xmajorgrids,
scatter/classes={%
    a={mark=triangle*,draw=green, fill=green, mark size=7},
    c={mark=triangle*,draw=pink, fill=pink, mark size=7},
    b={mark=triangle*,draw=blue, fill=blue, mark size=7}},
],
\addplot[scatter,only marks,%
    scatter src=explicit symbolic]%
table[meta=label] {
x y label
0 78.81 a
    };
\addlegendentry{FT}
\addplot[scatter,only marks,%
    scatter src=explicit symbolic]%
table[meta=label] {
x y label
1 66.23 c
    };
\addlegendentry{ANP}
\addplot[scatter,only marks,%
    scatter src=explicit symbolic]%
table[meta=label] {
x y label
0.5 79.60 b
    };
\addlegendentry{ANP+FT}

\path [fill=steelblue31119180, fill opacity=0.5]
(axis cs:0,79.4862347434693)
--(axis cs:0,78.2215210488258)
--(axis cs:0.0526315789473684,78.3584816872314)
--(axis cs:0.105263157894737,78.5897328366071)
--(axis cs:0.157894736842105,78.7975880037473)
--(axis cs:0.210526315789474,78.8073696250663)
--(axis cs:0.263157894736842,79.0198946743595)
--(axis cs:0.315789473684211,79.1985616474256)
--(axis cs:0.368421052631579,79.3192912705022)
--(axis cs:0.421052631578947,79.2659226629036)
--(axis cs:0.473684210526316,78.9395989532984)
--(axis cs:0.5,78.8798634130995)
--(axis cs:0.526315789473684,78.612544100728)
--(axis cs:0.578947368421053,78.0400837424087)
--(axis cs:0.631578947368421,77.3848802199437)
--(axis cs:0.684210526315789,76.4169594768555)
--(axis cs:0.736842105263158,75.0856197898316)
--(axis cs:0.789473684210526,73.3896748901421)
--(axis cs:0.842105263157895,71.5529195254342)
--(axis cs:0.894736842105263,69.4724874257364)
--(axis cs:0.947368421052632,67.495488186473)
--(axis cs:1,65.1857417989398)
--(axis cs:1,67.2081673001743)
--(axis cs:1,67.2081673001743)
--(axis cs:0.947368421052632,69.4707018186527)
--(axis cs:0.894736842105263,71.3774879174964)
--(axis cs:0.842105263157895,73.3822923247214)
--(axis cs:0.789473684210526,75.1432739407181)
--(axis cs:0.736842105263158,76.7270472269244)
--(axis cs:0.684210526315789,77.9836490399985)
--(axis cs:0.631578947368421,78.9232211318813)
--(axis cs:0.578947368421053,79.5369820443023)
--(axis cs:0.526315789473684,80.0654456050917)
--(axis cs:0.5,80.3174287175828)
--(axis cs:0.473684210526316,80.3659846731693)
--(axis cs:0.421052631578947,80.6655471696786)
--(axis cs:0.368421052631579,80.688715832576)
--(axis cs:0.315789473684211,80.5444863503279)
--(axis cs:0.263157894736842,80.3540214069465)
--(axis cs:0.210526315789474,80.1301771280385)
--(axis cs:0.157894736842105,80.0871855070261)
--(axis cs:0.105263157894737,79.8731307182526)
--(axis cs:0.0526315789473684,79.6359404037964)
--(axis cs:0,79.4862347434693)
--cycle;

\addplot [semithick, steelblue31119180, mark=*, mark size=3, mark options={solid}]
table {%
0 78.8538778961475
0.0526315789473684 78.9972110455139
0.105263157894737 79.2314317774298
0.157894736842105 79.4423867553867
0.210526315789474 79.4687733765524
0.263157894736842 79.686958040653
0.315789473684211 79.8715239988768
0.368421052631579 80.0040035515391
0.421052631578947 79.9657349162911
0.473684210526316 79.6527918132338
0.5 79.5986460653412
0.526315789473684 79.3389948529098
0.578947368421053 78.7885328933555
0.631578947368421 78.1540506759125
0.684210526315789 77.200304258427
0.736842105263158 75.906333508378
0.789473684210526 74.2664744154301
0.842105263157895 72.4676059250778
0.894736842105263 70.4249876716164
0.947368421052632 68.4830950025628
1 66.1969545495571
};

\addplot[scatter,only marks,%
    scatter src=explicit symbolic, forget plot]%
table[meta=label] {
x y label
0 78.81 a
    };
\addplot[scatter,only marks,%
    scatter src=explicit symbolic, forget plot]%
table[meta=label] {
x y label
1 66.23 c
    };
\addplot[scatter,only marks,%
    scatter src=explicit symbolic, forget plot]%
table[meta=label] {
x y label
0.5 79.60 b
    };

\end{axis}

\end{tikzpicture}}
        \captionof{figure}{Impact of interpolation coefficient between FT and ANP}
        \label{fig:ablation_alpha}
    \end{minipage}
\end{figure}

We conduct an ablation analysis on each module of our proposed method in Table~\ref{tab:ablation_table}. Overall, averaging the negative prototypes (ANP) gives a more robust prototype for the negative class than using the negative classes separately (MNP). Furthermore, while transferring a single threshold $\bar{\lambda}$ from ImageNet1K (FT) is a competitive baseline, combining the adaptive threshold from ANP with the fixed threshold allows to further increase the performance. This highlights the importance of restraining the positive predictions in a certain region around the positive class.

In addition, we vary the interpolation coefficient between the fixed threshold and the adaptive in Figure~\ref{fig:ablation_alpha}. The plot shows that the optimal value lies around $0.42$ with a drop of performance for higher values than $0.6$. This confirms the effectiveness of blending the adaptive and fixed thresholds and that the choice of $\frac{1}{2}$ is a natural choice which yields good results without the introduction of additional hyperparameters.

\section{Conclusion}
In this paper, we consider the novel problem of zero-shot one-class classification. We propose a well performing method based on an LLM. We run extensive experiments to show the validity of our proposed method in hard cases and on both fine-grained and coarse tasks, which we believe reflects a realistic setting. 

We envision two main axes for improving over the proposed method. First, the fact that we rely on a threshold estimated from ImageNet1K might be problematic for application cases where the range of optimal thresholds would be very different. Second, the proposed method uses an LLM to generate candidates for visually confusing objects, which might be impractical in some settings (\emph{e.g.}, edge devices) or when one deals with very specific concepts.
\section*{References}
\medskip
  {\small
\bibliographystyle{unsrtnat}
\renewcommand{\bibsection}{}
\bibliography{egbib}}
\clearpage
\appendix



\section{iNaturalist benchmark}

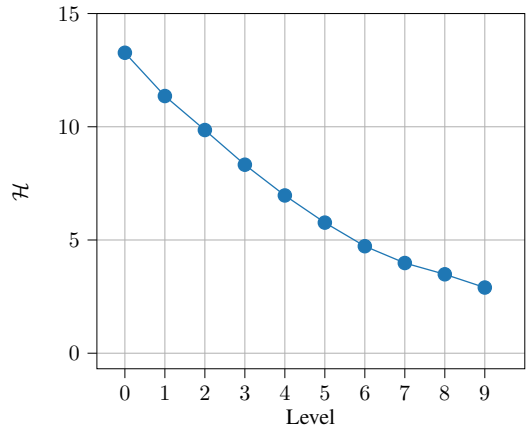
\begin{figure}[htbp!]
    \centering
    \resizebox{0.5\textwidth}{!}{
\begin{tikzpicture}

\definecolor{darkgray176}{RGB}{176,176,176}
\definecolor{steelblue31119180}{RGB}{31,119,180}

\begin{axis}[
tick align=outside,
tick pos=left,
x grid style={darkgray176},
xlabel={Level},
xmajorgrids,
xmin=-0.7, xmax=10, 
xtick style={color=black},
y grid style={darkgray176},
ylabel={$\mathcal{H}$},
ymajorgrids,
ymin=-0.686438542477013, ymax=15,
ytick style={color=black},
xtick={0,1,2,3,4,5,6,7,8,9}
]
\addplot [semithick, steelblue31119180, mark=*, mark size=3, mark options={solid}]
table {%
0 13.2682495719928
1 11.3571721578481
2 9.85595016783374
3 8.32413775761558
4 6.96664420306323
5 5.7672486154608
6 4.72544328911013
7 3.98312896213576
8 3.4850448183548
9 2.90061181960795
};
\end{axis}

\end{tikzpicture}}
    \caption{Entropy of the sampling in the hierarchical taxonomy of iNaturalist.}
    \label{fig:entropy}
\end{figure}

In this section, we describe the obtained sampling in iNaturalist. We report in Figure~\ref{fig:entropy} the entropy over all possible nodes in the taxonomy per level. At level 0, the possible node candidates are all the leaves. As the level increases, the number of possible nodes decreases. We compute the entropy as follows: 
\begin{align}
    \mathcal{H} = -\sum_{i=1}^{N} p(v_i) \log_2(p(v_i))
\end{align}
where $N$ represents the number of distinct nodes at each level $L$ of the taxonomy tree, $p(v_i)$ is the probability of selecting a node $v_i$ at level $L$. 

\begin{figure}[h!]
    \centering
    \subfloat[\centering Level 0, class name is ``Mugwort'']{\includegraphics[scale=0.17]{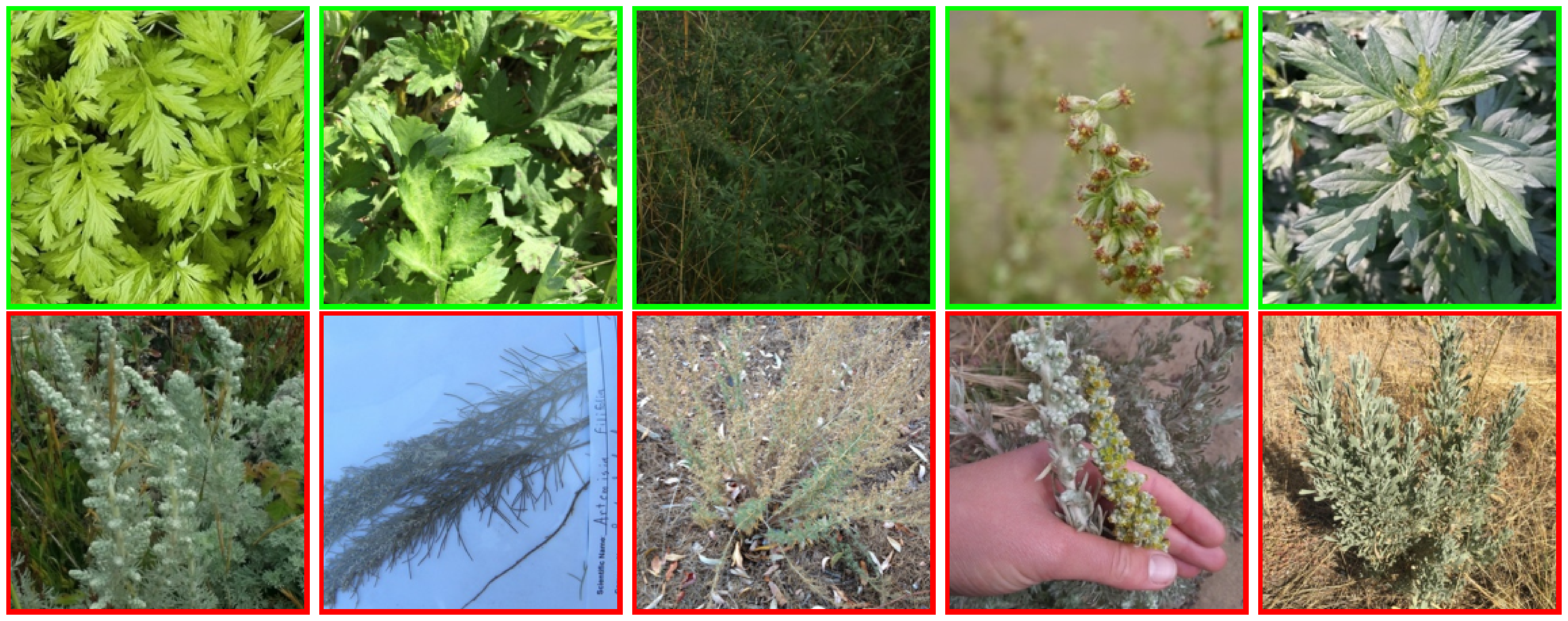}}\hfil 
    \subfloat[\centering Level 5, class name is ``Vertebrates'']{\includegraphics[scale=0.17]{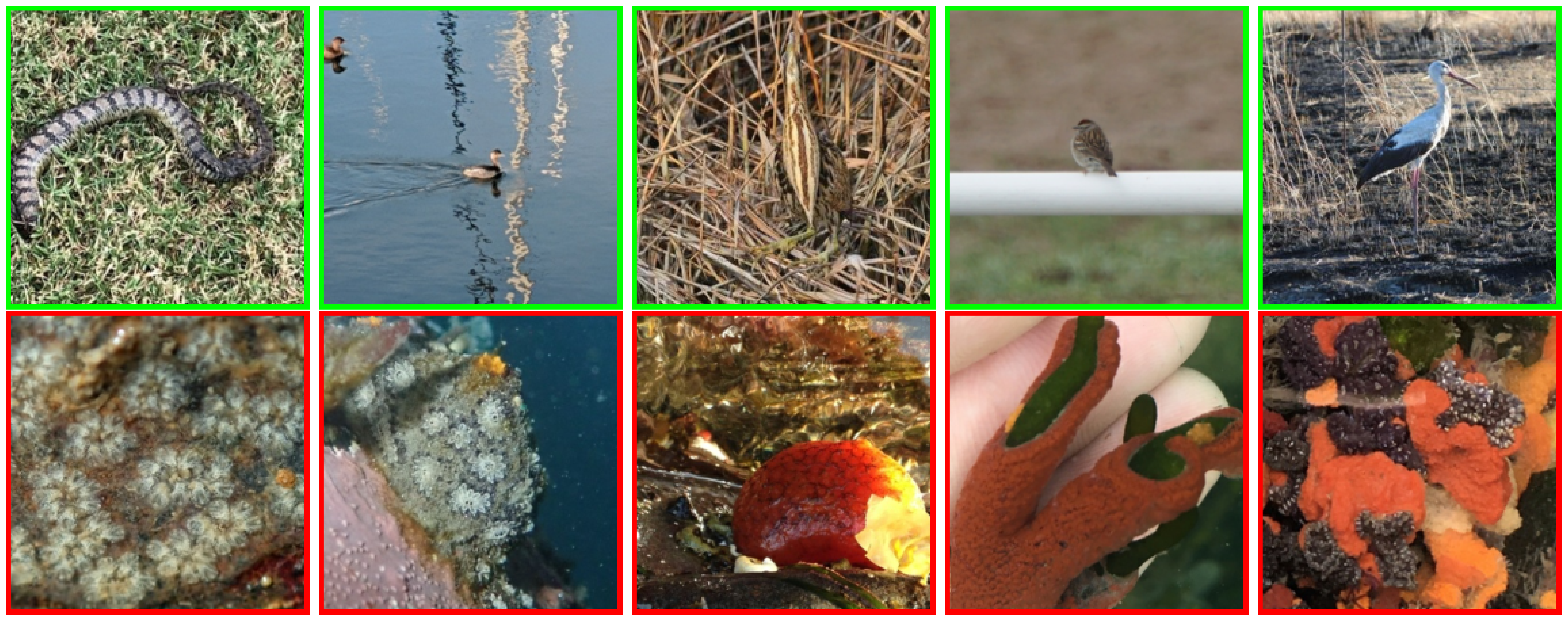}}\hfil 
        \subfloat[\centering Level 9, class name is ``Plants'']{\includegraphics[scale=0.17]{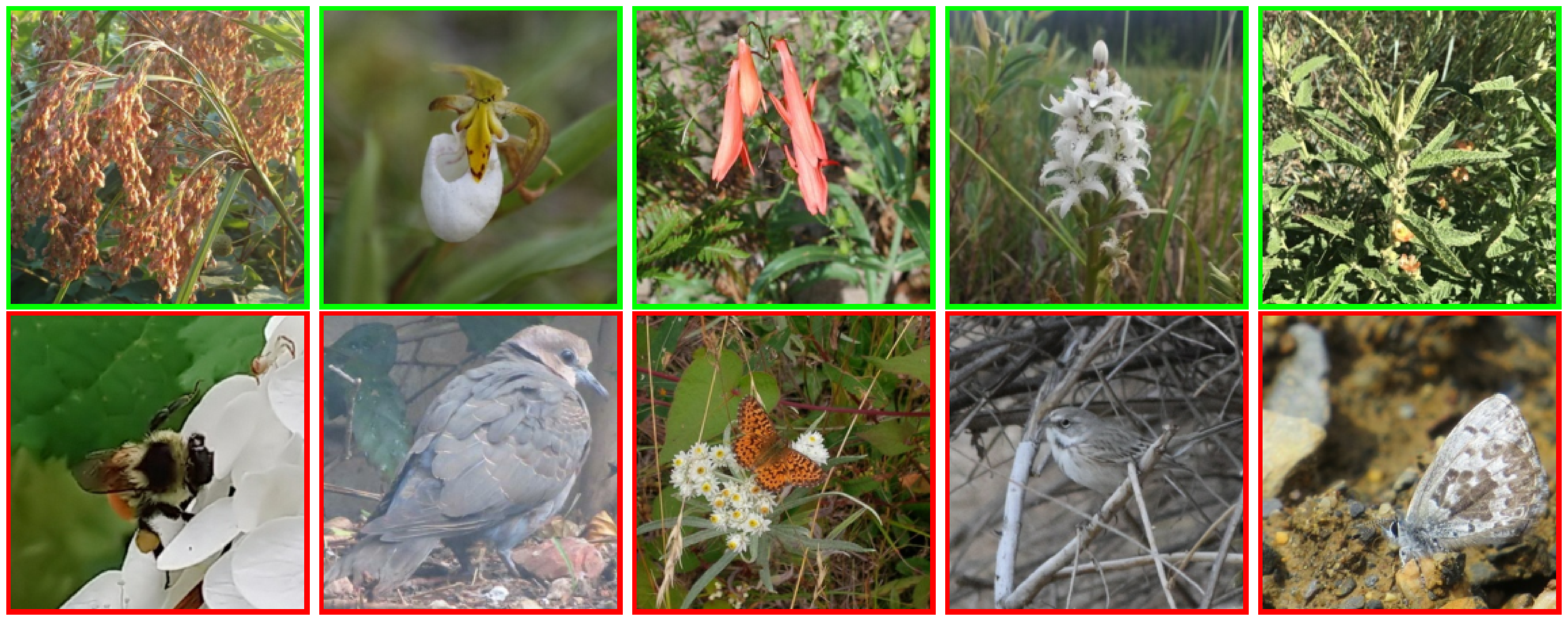}}\hfil 
    \caption{Examples of query samples from the proposed hierarchical sampling for iNaturalist. Positive query samples are highlighted in green while negative query samples are framed in red.}
    \label{fig:examples_inat}
\end{figure}

Furthermore, we provide some examples of different tasks sampled at different levels in Figure~\ref{fig:examples_inat}. Lower levels contain hard tasks with close species, while higher levels contain broader concepts and make easier tasks.

\FloatBarrier
\section{Quality of visually confusing classes}
In this experiment, we investigate the quality of the confusing classes obtained using the Large Language Model. Instead of using visually confusing classes from the LLM, we use the existing groundtruth classes available in the original datasets. For each class label, we rank the other classes based on their cosine similarity to the target class and retain the 10 closest ones. For iNaturalist dataset, we restrict these classes to the neighbouring classes, \emph{i.e.}, the ones sharing a common ancestor with the target class. We keep the same methodology (ANP+FT) and only change the LLM based confusing classes with the groundtruth classes.

We report the results in Table~\ref{tab:gt_classes} where we observe that our LLM based method is not far from the groundtruth based method. This hints that the LLM captures well visually confusing classes. With the noticeable exception of iNaturalist where the LLM performs better. The first possible explanation is that iNaturalist contains close concepts which can be better captured with the LLM. The second explanation is that the LLM generates common names for the visually confusing classes of iNaturalist which work better for CLIP.

\renewcommand{\arraystretch}{1.2}

\begin{table}[h!]
  \caption{Groundtruth classes \emph{vs.} LLM prompts for negative prototypes, macro F1 score.
  }
   \label{tab:gt_classes}
  \centering
  \resizebox{\textwidth}{!}{\begin{tabular}{@{}c|c|c|c|c|c|c|c@{}}
    \hline 
\rule{0pt}{15pt} 
Method & iNat	& SUN	& Food	& Textures	& Pets & EuroSAT	& Average \\
\hline

 LLM&$\mathbf{60.32\pm0.99}$&$94.19\pm0.38$&${94.18\pm0.20}$&${78.21\pm0.88}$&${90.97\pm0.88}$&$59.72\pm0.98$&${79.60}$\\
 
Groundtruth &$54.51\pm0.98$&$\mathbf{95.39\pm0.30}$&$\mathbf{94.25\pm0.22}$&$\mathbf{80.37\pm0.83}$&$\mathbf{91.40\pm0.80}$&$\mathbf{65.33\pm0.99}$&$\mathbf{80.21}$\\
\hline
\end{tabular}}
 
\end{table}

\section{Examples of visually confusing classes}
In this section, we provide some examples for the visually confusing objects generated using the LLM. 
We highlight in blue classes that are present in the original dataset, which were captured by the LLM. We observed on a large number of examples that the negative classes given by the LLM on EuroSAT and Textures often do not match existing classes in these two datasets, which could explain the lower performance compared to other datasets.
\begin{itemize}
\item \textbf{(iNat)} Birds: \textcolor{blue}{Bats}, \textcolor{blue}{Flying Squirrels}, \textcolor{blue}{Butterflies}, \textcolor{blue}{Moths}, \textcolor{blue}{Dragonflies}, \textcolor{blue}{Flying Fish}, \textcolor{blue}{Flying Lemurs}, \textcolor{blue}{Flying Frogs}, \textcolor{blue}{Flying Snakes}, Pterosaurs.
\item \textbf{(SUN)} Abbey: \textcolor{blue}{Cathedral}, \textcolor{blue}{Church}, \textcolor{blue}{Monastery}, \textcolor{blue}{Synagogue}, \textcolor{blue}{Palace}, \textcolor{blue}{Temple}, \textcolor{blue}{Mosque}, \textcolor{blue}{Castle}, \textcolor{blue}{Mansion}, Chapel.
\item \textbf{(Pets)} Pug: \textcolor{blue}{French Bulldog}, \textcolor{blue}{Boxer}, \textcolor{blue}{Shih Tzu}, \textcolor{blue}{Cavalier King Charles Spaniel}, Boston Terrier, Bulldog, Bullmastiff, Staffordshire Bull Terrier, Lhasa Apso, Pekingese.
\item \textbf{(Food)} Churros: \textcolor{blue}{French Fries}, \textcolor{blue}{Spring Rolls}, \textcolor{blue}{Donuts}, Pretzels, Breadsticks, Cinnamon Sticks, Fries, Mozzarella Sticks, Cinnamon Rolls, Taquitos.
\item \textbf{(Textures)} Bubbly: \textcolor{blue}{Pitted}, \textcolor{blue}{Porous}, Foamy, Frothy, Spongy, Puffy, Effervescent, Airy, Popcorn-like, Blisteblue.
\item \textbf{(EuroSAT)} River: Canal, Stream, Creek, Brook, Estuary, Fjord, Strait, Channel, Bayou, Inlet.
\end{itemize}


    


\section{Additional ablations}

\begin{figure}[htbp]
    \centering
    \resizebox{0.5\textwidth}{!}{
\begin{tikzpicture}

\definecolor{darkgray176}{RGB}{176,176,176}
\definecolor{steelblue31119180}{RGB}{31,119,180}

\begin{axis}[
height=8cm, 
width=8cm,
tick align=outside,
tick pos=left,
x grid style={darkgray176},
xlabel={Number of negative prototypes (K)},
xmin=0, xmax=31.25,
xtick style={color=black},
y grid style={darkgray176},
ylabel={macro F1 score (\%)},
ymin=60, ymax=85,
ymajorgrids,
xmajorgrids,
ytick style={color=black},
xtick={1,5,10,15,20,25,30}
]

\path [fill=steelblue31119180, fill opacity=0.5]
(axis cs:1,75.91)
--(axis cs:1,74.44)
--(axis cs:5,78.37)
--(axis cs:10,78.97)
--(axis cs:20,77.92)
--(axis cs:30,77.31)
--(axis cs:30,78.59)
--(axis cs:30,78.59)
--(axis cs:20,79.20)
--(axis cs:10,80.22)
--(axis cs:5,79.66)
--(axis cs:1,75.91)
--cycle;

\addplot [semithick, steelblue31119180, mark=*, mark size=3, mark options={solid}]
table {%
1 75.18
5 79.01
10 79.60
20 78.56
30 77.95
};
\end{axis}

\end{tikzpicture}}
    \caption{Impact of varying the number of negative prototypes.}
    \label{fig:nb_negative_prototypes}
\end{figure}
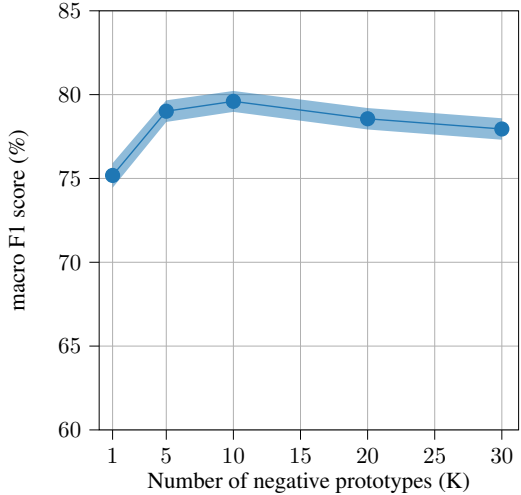

In Figure~\ref{fig:nb_negative_prototypes}, we vary the number of visually confusing objects from the LLM and report the average performance. The figure shows that the best performance is at $10$ negative prototypes. Using too few negative prototypes ($1$ or $5$) performs worse, which is expected since it becomes more prone to LLM mistakes and may not be enough to capture the target class boundaries properly. On the other hand, there is a diminishing return effect with prompting for too many negative classes since the LLM might output redundant negative classes leading to biases. 

\FloatBarrier
\section{Usage of common names vs scientific names}
So far in our experiments for iNaturalist, we have used common names instead of the official scientific names provided by the iNaturalist dataset. In this experiment, we rerun the comparison between our method and CLIPN using the scientific names and report the results in Table~\ref{tab:common_names}. Using the common names instead of the original scientific names significantly improves the performance across all methods. This is expected as CLIP has not been trained on scientific names which reduces its zero-shot capabilities~\cite{scientificnamesinaturalist}.


\begin{table}[htbp!]
  \caption{Comparison between common and scientific names on iNaturalist.
  }
   \label{tab:common_names}
  \centering
  \resizebox{0.5\textwidth}{!}{
\begin{tabular}{@{}c|c|c@{}}
    \toprule 
 Method &	Common names & Scientific names	\\
\hline 
CLIPN~\cite{clipn}+FT &$\mathbf{50.03\pm1.11}$&$45.61\pm0.91$\\
ANP+FT &$\mathbf{60.32\pm0.99}$&$54.43\pm1.05$ \\
\bottomrule
\end{tabular}
}
\end{table}

\section{Accuracy and AUC scores}
In this section, we report additional metrics for our main experiments. We first report the accuracy in Table~\ref{tab:accuracies}. Note that for these experiments, the ratio of positive samples is $r=0.5$. Overall, we get similar results to the ones obtained using the macro F1 score.

\begin{table}[h!]
  \caption{Accuracy score for multiple datasets.
  }
   \label{tab:accuracies}
  \centering
  \resizebox{\textwidth}{!}{\begin{tabular}{@{}c|cccccc|c@{}}
    \toprule 
Method &	iNat	& SUN	& Food	& Textures	& Pets & EuroSAT	& Average \\
\midrule
CLIPN~\cite{clipn}&$54.25\pm0.47$&$83.03\pm0.48$&$72.15\pm0.60$&$60.24\pm0.43$&$69.81\pm0.62$&$\mathbf{69.18\pm0.75}$&$68.11$\\
ZOC\cite{zoc} &$60.24\pm0.82$&$93.68\pm0.31$&$92.56\pm0.24$&$75.22\pm0.59$&$88.75\pm0.57$&${68.63\pm0.91}$&$79.84$\\
{ANP+FT} (ours) & $\mathbf{64.61\pm0.80}$&$\mathbf{94.29\pm0.35}$&$\mathbf{94.20\pm0.20}$&$\mathbf{79.26\pm0.74}$&$\mathbf{91.71\pm0.71}$&${63.93\pm0.79}$&$\mathbf{81.33}$\\
\bottomrule
\end{tabular}}
\end{table}

Furthermore, we report in Table~\ref{tab:aucs} the AUC score. Although being threshold invariant, the AUC reports shows how well the positive and negative samples are separated. The table shows that our method obtains on average competitive AUC scores. Our method outperforms alternatives on the iNaturalist dataset, where AUC scores are relatively low due to the difficulty of this setting. For other datasets, the different methods yield overall a good performance with an AUC score up to $99\%$ on SUN and Food. However, a high AUC score does not translate directly to a good F1 score or accuracy, as it is still important to select an appropriate threshold. Note that our main contribution is to propose a discriminative method, which rejects the negative samples without a need of calibrating the threshold on validation samples drawn from the task. 

\begin{table}[h!]
  \caption{AUC score for multiple datasets.
  }
   \label{tab:aucs}
  \centering
  \resizebox{\linewidth}{!}{\begin{tabular}{@{}c|cccccc|c@{}}
    \toprule 
Method &	iNat	& SUN	& Food	& Textures	& Pets & EuroSAT	& Average \\
\midrule
CLIPN~\cite{clipn}&$66.39\pm1.17$&$99.19\pm0.08$&$98.32\pm0.08$&$86.31\pm0.82$&$96.49\pm0.73$&$87.18\pm0.56$&$88.98$\\
ZOC\cite{zoc} &$69.28\pm1.10$&$\mathbf{99.21\pm0.10}$&$\mathbf{99.08\pm0.06}$&$\mathbf{90.32\pm0.57}$&$97.37\pm0.60$&$79.16\pm0.89$&$\mathbf{89.07}$\\
{ANP+FT} (ours)&$\mathbf{71.35\pm1.10}$&$98.76\pm0.20$&$98.76\pm0.08$&${86.56\pm0.96}$&$94.29\pm1.19$&${79.44\pm0.83}$&${88.19}$\\
\bottomrule
\end{tabular}}
\end{table}

\FloatBarrier





\section{How do we perform with comparison to visual shot-based methods?}
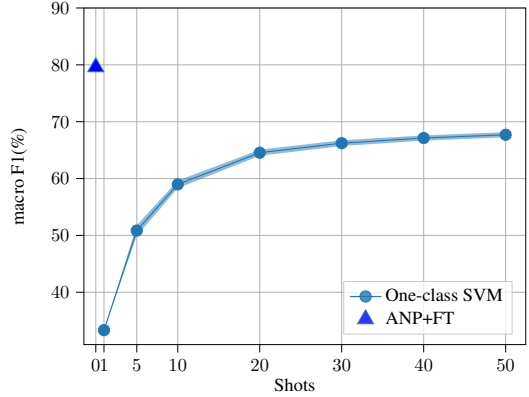
\begin{figure}[htbp!]
    \centering
    \resizebox{0.5\linewidth}{!}{
\begin{tikzpicture}

\definecolor{darkgray176}{RGB}{176,176,176}
\definecolor{green}{RGB}{0,128,0}
\definecolor{lightgray204}{RGB}{204,204,204}
\definecolor{pink}{RGB}{255,192,203}
\definecolor{steelblue31119180}{RGB}{31,119,180}

\begin{axis}[
height=8cm, 
width=10cm,
legend cell align={left},
legend style={
  fill opacity=0.8,
  draw opacity=1,
  text opacity=1,
  at={(0.97,0.03)},
  anchor=south east,
  draw=lightgray204
},
tick align=outside,
tick pos=left,
x grid style={darkgray176},
xlabel={Shots \textcolor{white}{\%}},
xmin=-1.45, xmax=52.45,
xtick style={color=black},
y grid style={darkgray176},
ylabel={macro F1(\%)},
ymin=30.791, ymax=90,
ytick style={color=black},
xtick={0,1,5,10,20,30,40,50},
ymajorgrids,
xmajorgrids,
scatter/classes={%
    b={mark=triangle*,draw=steelblue31119180, fill=blue, mark size=5},
    c={mark=triangle*,draw=violet, fill=violet, mark size=5}}
], 

\path [fill=steelblue31119180, fill opacity=0.5]
(axis cs:1,33.3333333333333)
--(axis cs:1,33.3333333333333)
--(axis cs:5,49.908407461731)
--(axis cs:10,58.3032013080783)
--(axis cs:20,64.0059509489384)
--(axis cs:30,65.7107282520685)
--(axis cs:40,66.6737970661188)
--(axis cs:50,67.2447780071698)
--(axis cs:50,68.1514447827568)
--(axis cs:50,68.1514447827568)
--(axis cs:40,67.5895730708985)
--(axis cs:30,66.7373743527836)
--(axis cs:20,65.0795461206946)
--(axis cs:10,59.6459498870599)
--(axis cs:5,51.7886005283573)
--(axis cs:1,33.3333333333333)
--cycle;

\addplot [semithick, steelblue31119180, mark=*, mark size=3, mark options={solid}]
table {%
1 33.3333333333333
5 50.8485039950441
10 58.9745755975691
20 64.5427485348165
30 66.2240513024261
40 67.1316850685086
50 67.6981113949633
};
\addlegendentry{One-class SVM}

\addplot[scatter,only marks,%
    scatter src=explicit symbolic]%
table[meta=label] {
x y label
0 79.60 b
    };
\addlegendentry{ANP+FT}
\end{axis}
\end{tikzpicture}}
    \caption{Average macro F1 score for different methods for different shots.}
    \label{fig:shots}
\end{figure}

We sample some shots for each task and train a one-class SVM classifier~\cite{oneclasssvm} on top of the visual features extracted from CLIP using its visual encoder. This classifier has been widely used for one-class classification~\cite{oneclasssurvey,drocc}. Figure~\ref{fig:shots} shows that text has strong representations since our methods outperform the one-class SVM even with up to 50 shots. Furthermore, given the hard nature of the sampled tasks, the SVM classifier performs poorly in the very low shot regime (\emph{i.e.}, $\leq 10$ shots).

\section{What does the boundary decision looks like for different methods?}
\begin{figure}[htbp!]
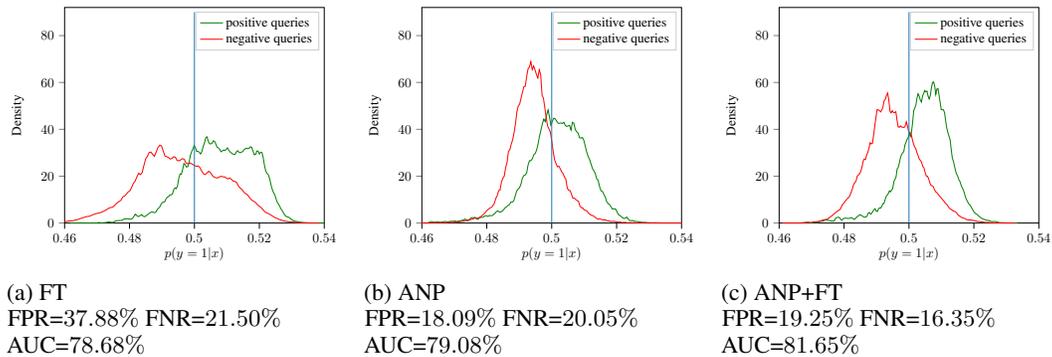

    \centering
    \begin{subfigure}{0.32\linewidth}
        \centering
        \resizebox{\linewidth}{!}{\input{plots/decision_boundaries/decision_boundaryFT}}
        \caption{FT\\FPR=$37.88\%$ FNR=$21.50\%$\\AUC=$78.68\%$}
    \end{subfigure}
    \hfill
    \begin{subfigure}{0.32\linewidth}
        \centering
        \resizebox{\linewidth}{!}{\input{plots/decision_boundaries/decision_boundaryANP}}
        \caption{ANP\\FPR=$18.09\%$ FNR=$20.05\%$\\AUC=$79.08\%$}
    \end{subfigure}
    \hfill
    \begin{subfigure}{0.32\linewidth}
        \centering
        \resizebox{\linewidth}{!}{\input{plots/decision_boundaries/decision_boundaryANPS}}
        \caption{ANP+FT\\FPR=$19.25\%$ FNR=$16.35\%$\\AUC=$81.65\%$}
    \end{subfigure}
    
    \caption{Density of probability outputs of each classifier across all datasets.}
    \label{fig:decision_boundary}
\end{figure}

In order to better understand the behaviour of each method, we plot in Figure~\ref{fig:decision_boundary} the decision boundaries for $200$ runs per dataset and per level for iNaturalist. Using a sigmoid function, we compute for each method the probability of accepting samples up to a certain scaling temperature. First, we observe that for the Fixed Threshold method (FT) the decision boundary is not properly calibrated, which is expected since the threshold from ImageNet1K is not appropriate for all tasks as shown in the main paper. Second, the separation between the positives and negatives queries is better when using ANP since the negative queries tend to get closer to the negative protypes. Finally, when combining the adaptive and fixed thresholding (ANP+FT), the separation between the positive and negative queries is further improved as hinted with a higher AUC score. We can also observe that the right tail of the negative distribution is smaller compared to ANP since we remove outliers by restricting the positive predictions.

\end{document}